# Texture image retrieval using a classification and contourlet-based features

**Asal Rouhafzay, Nadia Baaziz, Mohand Said Allili**

Département d'informatique et d'ingénierie, Université du Québec en Outaouais, Gatineau, Canada, Nadia.Baaziz@uqo.ca

**Abstract –** In this paper, we propose a new framework for improving Content Based Image Retrieval (CBIR) for texture images. This is achieved by using a new image representation based on the RCT-Plus transform which is a novel variant of the Redundant Contourlet transform that extracts a richer directional information in the image. Moreover, the process of image search is improved through a learning-based approach where the images of the database are classified using an adapted similarity metric to the statistical modeling of the RCT-Plus transform. A query is then first classified to select the best texture class after which the retained class images are ranked to select top ones. By this, we have achieved significant improvements in the retrieval rates compared to previous CBIR schemes.

**Keywords:** Contourlet transform; supervised learning; feature extraction; texture retrieval;

## *1 Introduction*

Content-based image retrieval (CBIR) is an important problem for browsing and retrieving images from large databases of visual media. In particular, CBIR for texture images has received the focus of several research work since it is essential for several applications such as remote sensing, industrial inspection, quality control, medical imaging, to name a few. However, there are remaining challenges for achieving accurate and computationally efficient texture retrieval mainly due to the cheer number of texture classes and the variability inside each texture class [1].

Usually, the performance of CBIR systems depends largely on the way the images are represented to facilitate their indexation and the type of similarity measurement used to compare images. In fact, a typical image retrieval system will be composed of two main steps: 1) feature extraction and modeling which transforms an image into a compact feature vector, 2) Similarity measurement (SM), which is used to compare images and return the closest ones to the query image. Several methods have been proposed to improve these two steps. In terms of texture features, image representation approaches have used spectral, multiscale, statistical, geometric methods or their combination [2], [3]. Among the most successful methods are statistical methods combined with multiscale methods which have been proposed recently. For example, modeling wavelet distribution using the formalism of the generalized Gaussian density (GGD) has been investigated in several research works. An extension of this method using the contourlet transform and the formalism of Gaussian mixtures have allowed to sensibly improve the performance of texture retrieval [4].

In terms of similarity measurement, statistical distances have been extensively used for comparing images. On the one hand, when images are represented in vector spaces, usual norms can be employed [1]. On the other hand, when images are described in terms of random processes, statistical divergences have been used [3]. Despite their performance, these models can have limitation to handle the huge number of classes and variability within each class. Moreover, they require comparing each search query with the whole database images which is not practical for retrieval in large scale databases. Recently, machine learning techniques have been used to improve texture classification. These methods have been, notably motivated by applications such as remote sensing and industrial inspection. In particular, methods have been proposed for texture discrimination using K-nearest neighbor (kNN), support vector machines (SVM) and naïve Bayes (NB). However, these methods are more adapted to application scenarios where the number of classes is very limited and the algorithms are not confronted to handle large scale data.

In this paper, we propose a new framework integrating learning-based texture representation and retrieval. Building on our past work [5], we propose to extract richer features from texture images by augmenting the directional selectivity of the Redundant Contourlet transform, which enables accurate description of image structures describing the texture content. Using the GGD signature to represent the distribution of such a transform, we develop the *symmetric Kullbak-Leibler divergence* as a similarity measure for the purpose of query to image comparison and ranking. Moreover, we propose a CBIR system with machine learning approach (ML-CBIR) that performs search in two phases. In the first phase, a query is classified according to pre-existing texture classes in the database. In the second phase, the top class is returned and each image in the returned class is ranked according to its similarity to the query image.

This system allows not only to enhance the retrieval score but also to make it computationally efficient avoiding the requirement of comparing a query with all the images of the database. We show through several experiments using standard datasets that our approach outperforms recent state-of-art methods.

In section 2, we present our CBIR scheme for texture images and some related works. Section 3 is dedicated to Contourlet modeling, feature extraction, indexing and similarity measurement. In section 4, we provide the experiments and results for the evaluation. We end the paper with a conclusion and future perspectives.

## *2 Machine learning in the CBIR scheme*

In traditional CBIR systems, the general architecture is organized into two separate phases: The *Offline* phase and the *Online* phase. The *Offline* phase is mainly dedicated to the extraction of a compact feature vector from each image in the database to build the database index (feature dataset). The *Online* phase operates a database search to find and retrieve relevant images to the user query. Similarity measurements, using a distance metric, are first calculated between the query feature vector and every feature vector in the database index. Next, The *N* smaller similarity measures are selected as *TopN* matches and the corresponding images are retrieved and presented in a ranked order to the user as being the most relevant to the query image (*TopN* retrieved images).

### 2.1 Related work

Taking advantage from supervised machine learning algorithms for classification purpose is a possible approach in a CBIR system. In the recent years, several works with have been proposed by the researchers and most of them combined many extracted features to perform image retrieval. In [6], Sugamya et al. proposed a CBIR system which combined three low-level feature extraction techniques: color-correlogram for color feature extraction, wavelet transformation for extracting shape features and Gabor wavelet for texture feature extraction. Further, a support vector machine classifier was trained to classify the features of a query image and distinguish between relevant and irrelevant images accordingly. This method led to better performance than the traditional method of image retrieval and achieved results were found encouraging in terms of image classification accuracy. In 2017, Kaur et al. [7] proposed a CBIR system based on statistical image features, such as skewness, kurtosis and standard deviation, which were extracted from the probability color histogram of database images. The training phase was performed on Corel-1000 database using one of the three classifiers: artificial neural network, naïve Bayes and neuro-fuzzy network. These retrieval approaches were compared on the basis of precision and error rate. Neuro-fuzzy classifier outperformed other techniques on the experimented dataset. In 2018 Toroitich et al. [8] presented their approach in CBIR using the combination of color, texture features and kNN. An Euclidean distance metric was used to determine the nearest objects, thus resulting in the least number of images retrieved by the system. For kNN algorithm, different k values were tested to determine the best value for different classes of images. In 2019, Al-Qasemy et al. [9] have also used the kNN supervised learning to train statistical features (mean, standard deviation and skewness) which were extracted from either RGB or grayscale database images. Image retrieval efficiency was measured by means of precision, recall and f-measure. The proposed feature selection method resulted in high precision retrieval. In [10], Alrahha and Supreeti firstly proposed the local neighbor pattern (LNP) method for image feature extraction, which resulted in better retrieval recall when compared with other methods such as local binary pattern (LBP). Secondly, the CBIR system was improved in terms of accuracy using supervised machine learning such as SVM and kNN on three different image databases.

**2.2 Proposed CBIR framework**

A supervised machine learning approach (ML) is incorporated into the CBIR scheme for texture retrieval. The block diagram of the proposed ML-CBIR framework is depicted in Fig.1. During the *Offline* phase, the extracted texture feature vectors from the image database are also labeled to indicate their class membership (image category) according to pre-existing texture classes. In addition to feature extraction and feature vector labeling (index), a training procedure is conducted on the labeled feature vector dataset in order to build a classification model with high accuracy. Among well-known supervised learning algorithms for classification, one can find neural networks variants, support vector machines (SVM), k-nearest neighbors (kNN), decision trees and discriminant analysis [11], [12].

During the *Online* phase, the search and retrieval process to any given query image is performed through query image classification. Firstly, the trained classifier is applied to the given query feature vector in order to predict its class membership (class label). Next, all feature vectors from the predicted class are compared to the query feature vector using a similarity metric. The *N* smaller similarity indices are selected as the *TopN* matches. The corresponding images are retrieved in a ranked order and returned to the user as the most relevant texture images to the query (*TopN* retrieved images).

In addition to the incorporated ML approach in the proposed framework, the choice of a texture feature extraction methodology and adapted similarity metric play a key role in the efficiency of the texture retrieval process.

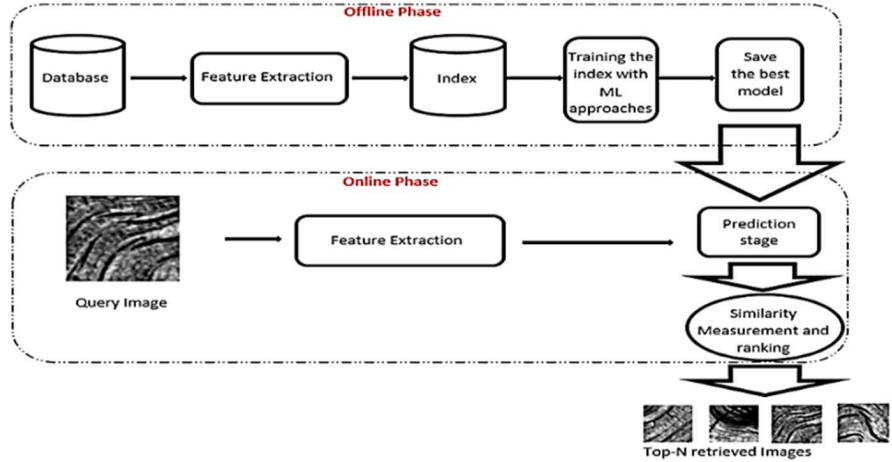

**Fig. 1.** Proposed framework of texture retrieval using Machine Learning approach (ML-CBIR).

## *3 Contourlet modeling and feature extraction*

Extracted features from the input image dataset are anticipated to contain relevant texture information in order to achieve powerful characteristic discrimination while ensuring compact data representation. We focus on the statistical modeling of texture sub-bands through multiscale image decomposition using RCT-Plus transform. In our previous work [5], this approach has proved itself as reasonably appropriate for feature extraction in a traditional CBIR framework.

**3.1 RCT-Plus decomposition scheme**

RCT-Plus is a new variant of the contourlet transform [13] in which augmented directional selectivity, redundancy and Gaussian filtering for multiscale decomposition are meant to enhance texture characterization and extract a richer directional information in the image [5]. RCT-Plus is created by two main stages, a multiscale decomposition followed by a directional sub-band decomposition. The first stage implements a Redundant Laplacian Pyramid (*RLP*). Mainly, a set of *L* Pseudo-Gaussian low-pass filters are used to filter input image to provide *L+1* equal-size sub-images: *L* detail sub-images (*RLP*$_1$) at scale level *l* and one low-pass image approximation $C_L$. All input *RLP* sub-images are of the same size as the input image since the redundant

Laplacian pyramid is not down-sampled. The second stage is a directional filter bank. For each *RLP* sub-image at scale level *l*, a *2-D* filter bank with $D_l$ orientations and critical sub-sampling is applied to obtain a number $D_l$ of directional sub-bands $\{C_{ld}; l = 1 ... L; d = 1 ... D_l\}$. The allowed values for $D_l$ are in the set $\{2, 4, 8...\}$ and can vary from one scale level to another. Fig. 2 shows an example of a 3-scale level RCT-Plus decomposition of a texture image resulting into 17 sub-bands.

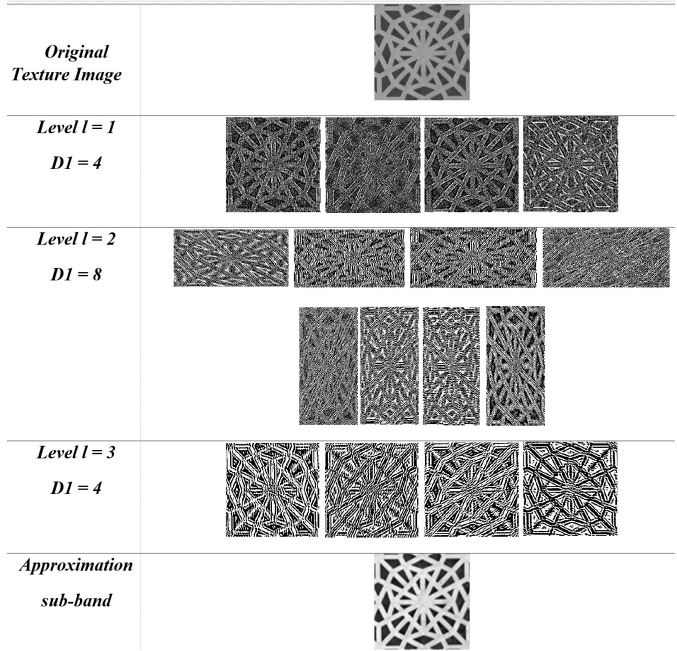

**Fig. 2.** RCT-Plus decomposition example for 3 scale levels *(L=3)*. The number of directional sub-bands at each scale level *l* is indicated by $D_l$.

### 3.2 Feature extraction and similarity measurement

Two modeling methodologies using generalized Gaussian distribution (GGD) and energy, respectively, are applied on the sub-bands of a multiscale image representation such as the RCT-Plus. Given a texture image *I* and its RCT-Plus decomposition yielding *L* scale levels and $D_l$ directional sub-bands at each scale level *l*, we have modeled the marginal distribution for each sub-band by a GGD function and then a model parameter estimator, namely, *Moment Matching estimator (MME)* or *Maximum Likelihood estimator (MLE)* is applied to estimate the scale and the shape parameters (α and β) of the fitting GGD. Once $\alpha$ and $\beta$ values for each sub-band are calculated, they are concatenated to form the texture feature vector $F_I$:

$$F_I = \{\alpha(l,d), \beta(l,d); \quad l=1...L; d=1...D_l\} \tag{1}$$

The *symmetric Kullback-Leibler Divergence* (KLD) is a statistical measure of how different two probability distributions over the same event space are. Once the KLD between all multi-scale sub-band pairs of two images are calculated, the summation of all KLD values is considered as the distance measure between two images. Consequently, similarity measurement between query image $Q$ and test image $T$ requires the computation of KLD measures between GGD probability distributions. It is formulated as the overall distance between all sub-band GGDs:

$$\delta_{TQ} = \sum_{l=1}^{L} \sum_{d=1}^{Dl} KLD\left(GGD_T^{ld}, GGD_Q^{ld}\right) \tag{2}$$

where $GGD_T$ and $GGD_Q$ represent the statistical models calculated on the images $T$ and $Q$ for each sub-band at the *l-th* scale level and the d-th direction, respectively. Note that Euclidean distance (ED) between the feature vectors $F_Q$ and $F_T$ is another alternative for similarity measurement between the images $T$ and $Q$.

We also define a multiscale energy-based approach for texture feature extraction which consists in calculating energy and characterizing its distribution through the RCT-Plus image sub-bands. It assumes that different texture patterns have different energy distribution in the space-frequency domain. This approach involves mainly the calculation of first and second order moments of the RCT-Plus directional sub-bands $\{C_{ld}; l = 1 ... L; d = 1 ... D_l\}$ and their concatenation into a feature vector $F_I$:

$$E(l,d) = \frac{1}{KM} \sum_{i=1}^{K} \sum_{j=1}^{M} |C_{ld}(i,j)| \tag{3}$$

$$F(l,d) = \sqrt{\frac{1}{KM} \sum_{i=1}^{K} \sum_{j=1}^{M} [C_{ld}(i,j)]^2} \tag{4}$$

$$F_I = \{E(l,d), F(l,d); \quad l = 1 ... L; \ d = 1...D_l\} \tag{5}$$

Note that K×M indicates the size of the considered sub-band $C_{ld}$. Similarity measurement is computed as the Euclidean distance (ED) between two compared feature vectors. The texture features that are derived from either GGD or Energy modeling approaches are compact vectors. Indeed, for an RCT-Plus with 3 scale levels and 8 directions at each scale level, including the approximation sub-band, the total number of feature vector components is 50 regardless the size of the original image I. Therefore, we define the feature extraction methods and similarity measurement that are being incorporated into the proposed ML-CBIR framework as follows:

GGD1: GGD modeling estimated by *MME*. KLD is used as a similarity metric.
GGD2: GGD modeling estimated by *MLE*. KLD is used as a similarity metric.
E: Energy modeling. Euclidean distance is used as a similarity metric.

## *4 Experiments and results*

**A. Dataset and evaluation criteria**

The proposed ML-CBIR framework has been tested on two texture databases: VisTex-40 and Kylberg-27 containing grayscale texture images from different scenes of daily life. The well-known MIT Vision Texture dataset [14] contains 40 distinct texture images, each of size 512×512. Each image was divided into 16 non-overlapping grayscale sub-images of size 128×128 thereby establishing a texture class. As a result, a database with 640 images organized into 40 texture classes was constructed and designated as VisTex-40 (see Fig. 3). We also extracted from the Kylberg Texture Dataset v. 1.0 [15], 27 distinct texture classes. Each class contains 40 grayscale images of size 512×512. As a result, a database with 1080 texture images organized into 27 textures classes was constructed for the experimentations and designated as Kylberg-27 (see Fig. 4). According to the work in [16], Kylberg is a more difficult database than the Vistex-40 for image retrieval purpose. To evaluate the efficiency of a supervised machine learning algorithm, one popular metric is *Accuracy* which is based on the prediction results obtained from applying the classifier to test data. The prediction results are formulated as four measures: true positives *tp*, false positives *fp*, true negatives *tn* and false negatives *fn*. The *Accuracy* measure is then given by the following ratio:

$$Accuracy = \frac{tp + tn}{tp + fp + tn + fn} \quad (6)$$

Cross-validation is a statistical approach widely used to determine how well a developed machine learning model can generalize the solution to new data. During the cross-validation, the whole training data set is randomly split into a number *n* of equal folds. The classifier model is then trained on *n-1* folds while the remaining fold is kept for testing purpose. This operation is iteratively repeated n times and each iteration deals with a newly selected testing fold. Consequently, we have n different values specifying the *Accuracy*. The average of the *n* obtained measures is then considered as the overall performance of the learned classifier model. This approach gives a more realistic evaluation of the model at expense of higher computational cost. Retrieval performance for each query image is measured in terms of the retrieval rate, which is calculated as the percentage of relevant images found among *N* retrieved images (*TopN* matches). Here, an image is considered relevant if it is part of the same class as the query. All retrieval results presented in this work are obtained by averaging the retrieval rates corresponding to all queries from the database *(AR %)*.

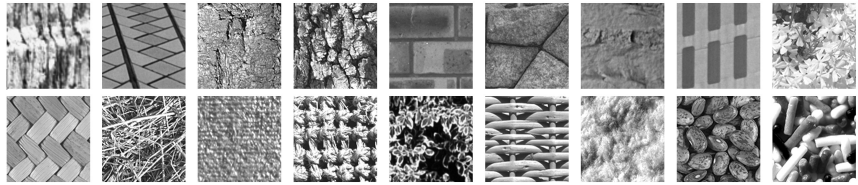

**Fig. 3.** Sample texture images from Vistex-40 database.

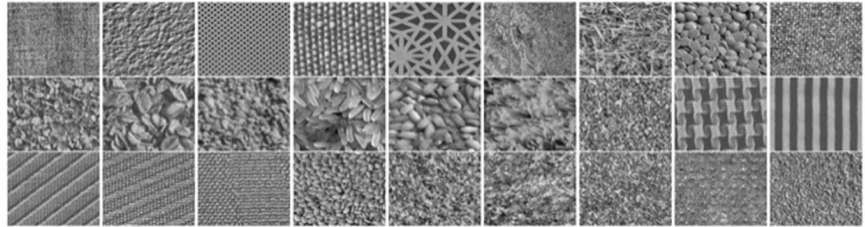

**Fig. 4.** Sample texture images from Kylberg-27 database.

**B. ML algorithms selection**

In order to study the accuracy of a selection of supervised ML algorithms such as kNN, SVM, Decision tree, linear discrimination and Quadratic discrimination, a learning phase with cross-validation is carried out for each algorithm using: *a)* A choice of parameters for each algorithm such as k value in kNN or kernel type in SVM, and *b)* Various data split scenarios in the *n-fold* cross-validation procedure (*3*-fold, *5*-fold …), and *c)* Vistex-40 and Kylberg-27 texture features derived from GGD1, GGD2 or E methods. We also decided to set the size of the RCT-Plus decomposition at 3 scale levels and 8 directions for all experiments, because this choice of parameters allowed us to achieve the best texture retrieval results in our previous work [5]. In most cases, kNN and SVM learning models were more accurate than Decision trees and Discriminant analysis models. Moreover, in most cases, kNN and SVM achieved high *Accuracy* levels when using GGD1 or GGD2 feature extraction methods. A sample of results on VisTex-40 database is shown in Table I where the best obtained *Accuracy* value is 99.69% for kNN (with *k=1*). Therefore, we consider that kNN and SVM algorithms are promising choices to be incorporated in the proposed ML-CBIR scheme.

**Table I:** *Accuracy* measures *(%)* of compared ML algorithms. The training features are from the whole VisTex-40 database.

| ML algorithms | Feature extraction methods using RCT-Plus | | |
|---|---|---|---|
| | GGD1 | GGD2 | E |
| kNN (k=1) | 99.69 | 98.00 | 96.95 |
| kNN (k=11) | 89.22 | 88.90 | 88.59 |
| SVM (Linear kernel) | 99.06 | 98.59 | 98.59 |
| SVM (Quadratic kernel) | 96.90 | 98.00 | 97.00 |
| SVM (Cubic kernel) | 96.90 | 97.50 | 96.10 |
| SVM(Gaussian) | 71.10 | 70.00 | 82.80 |
| Decision trees | 82.30 | 87.50 | 86.10 |
| Quadratic Discrimination | 96.12 | 94.87 | 91.42 |

**C. Retrieval results, comparison and discussion**

To evaluate the retrieval performance of the proposed ML-CBIR framework, we experimented separately the *Offline* and *Online* CBIR phases. Firstly, each classifier model (namely, kNN and SVM) is learned on a set of training features from VisTex-40. To this end, the training set (labelled index) corresponds to 600 images (15 per class) while 40 images are reserved for the *Online* phase test. Similarly, from Kylberg-27 database, 405 images are used as train data (15 images per class) while the remaining 675 images are kept as test data (25 images per class). The learned texture features are extracted using one of GGD1, GGD2 or E methods. Many other parameters are tuned such as the number of neighbors *(k=1, 2, 3, 4, 5, ...)* in the kNN algorithm, kernel type in SVM and the number of data partitions (*n*-fold) in the cross-validation procedure. Table II compares 6 different configurations of the ML-CBIR over the two databases. One can see that the trained classifiers (with *10*-fold cross validation) are of high quality since the *Accuracy* values exceed 97% for both kNN and linear SVM algorithms in conjunction with GGD1 or GGD2 feature extraction methods. Secondly, a CBIR *Online* phase testing is conducted. Each image of Vistex-40 and Kylberg-27 is considered as a query image and submitted to the ML-CBIR system. First, the trained classifier is used to predict the class membership (class label) of the submitted query image. Next, all images from the predicted class are retrieved and ranked according to the similarity measurement using either ED or KLD (for GGD1 and GGD2 features). Finally, the N first images are displayed as being the *N* most similar images to the query (*TopN* retrieved images), with *N* being equal to 15 at most.

The sample results reported in Table II also display a comparison of performance of 6 ML-CBIR configuration schemes in terms of Average retrieval rates *(AR%)* and the number of class membership predictions that are false. The considered ML algorithms are kNN (with *k=1*) and SVM (with Linear kernel). One can see that, all of the three feature extraction methods operating on RCT-Plus, namely GGD1, GGD2 and E, achieved average retrieval rates *(AR%)* that are higher than 99% in VisTex-40 and 93% in Kylberg-27 database. Regarding the impact of feature extraction on the retrieval performance, the GGD1 and GGD2 methods are more discriminative than energy-based method E. Moreover, SVM classification yields slightly better results than kNN classification. Indeed, the number of false class membership predictions (miss-classified queries) is very low (ranging from 0 to 5 out of 640 queries in the first database and from 22 to 71 out of 1080 queries in the second database). Fig. 5 presents an example where the kNN-CBIR system made false class membership predictions to 3 queries among 640 over Vistex-40. The first row of the table displays the submitted queries to the system while the second row illustrates a sample image of the false predicted class. In each case, the *TopN* retrieved images are non-relevant to the query. In Fig. 6 similar results are reported for an SVM-CBIR system over Kylberg-27.

**Table II:** Performance comparison of kNN-CBIR and SVM-CBIR schemes in terms of average retrieval rates *(AR %)* and *Accuracy* corresponding to a learning phase with *10*-fold cross-validations.

| ML algorithm: kNN (k =1) | | | | | | |
|---|---|---|---|---|---|---|
| | VisTex-40 | | | Kylberg-27 | | |
| | *AR%* | False predict | *Accuracy* | *AR%* | False predict | *Accuracy* |
| GGD1 | 99.69 | 2 | 97.33 | 97.69 | 25 | 99.26 |
| GGD2 | 99.53 | 3 | 97.33 | 97.31 | 29 | 98.76 |
| E | 99.22 | 5 | 96.66 | 93.43 | 71 | 94.81 |
| ML algorithm: SVM (Linear kernel) | | | | | | |
| | VisTex-40 | | | Kylberg-27 | | |
| | *AR%* | False predict | *Accuracy* | *AR%* | False predict | *Accuracy* |
| GGD1 | 99.86 | 1 | 98.67 | 97.96 | 22 | 99.26 |
| GGD2 | 99.86 | 1 | 98.67 | 97.59 | 26 | 99.51 |
| E | 99.69 | 2 | 98.17 | 93.43 | 71 | 98.52 |

A comparative study is done with traditional CBIR [5] and state-of-art method [16]. Table III and IV show that the proposed ML-CBIR framework outperforms traditional

CBIR which rely mainly on similarity measurement for searching and ranking feature vectors. Indeed, for both experimented databases, kNN-CBIR and SVM-CBIR yield significant increase of the retrieval rate (up to 32.64%) in comparison to traditional CBIR. More specifically, SVM-CBIR operating on RCT-Plus image features by means of GGD1 method gives the best performance out of all CBIR schemes discussed here (*AR%*=99.86 and *AR%*=97.96 for VisTex-40 and Kylberg-27 respectively). Another comparison is held with the state-of-art method in [16] referred to as Local directional peak valley binary patter (LDPVBP) for texture feature extraction and retrieval. The reported average retrieval rates are about 83.34% and 61.45% for VisTex and Kylberg databases respectively. This comparison confirms again the superiority of our proposed ML-CBIR framework.

**Table III:** Performance comparison of ML-CBIR schemes vs. traditional CBIR schemes over 640 images from Vistex-40 database.

| Feature extraction | *AR%* (kNN-CBIR) | *AR%* (Trad. CBIR) | Difference% |
|---|---|---|---|
| GGD1 | 99.69 | 77.15 | +22.54 |
| GGD2 | 99.53 | 75.24 | +24.29 |
| E | 99.22 | 71.72 | +27.50 |
| Feature extraction | *AR%* (SVM-CBIR) | *AR%* (Trad. CBIR) | Difference% |
| GGD1 | 99.86 | 77.15 | +22.71 |
| GGD2 | 99.86 | 75.24 | +24.62 |
| E | 99.69 | 71.72 | +27.97 |

## 5 Conclusion

In this paper, we proposed a new framework for improving Content Based Image Retrieval in databases of texture images. This is achieved by extracting compact and discriminative feature vectors using the statistical modeling of RCT-Plus image representation, a novel variant of the Redundant Contourlet transform which exhibits richer directional information in the image. Moreover, the process of image search and retrieval on a given database is improved through a supervised learning-based approach. Using either kNN or SVM algorithms, the query image is first classified according to pre-existing texture classes in the database. Then, a top class is returned and its contained images are ranked to select the $N$ relevant images to the query (*TopN*) according to an adapted similarity metric (KLD or ED). The experimental results on Vistex-40 and Kylberg-27 texture databases achieved a substantial increase up to +32% in terms of average retrieval rate *(AR%)* in comparison to traditional CBIR. In addition, the *Online* search phase in the proposed system is computationally efficient, thus avoiding the need to compare a query with all the images in the database.

As future work, the search and retrieval space can be extended by considering a multi-class membership prediction for the query classification process.

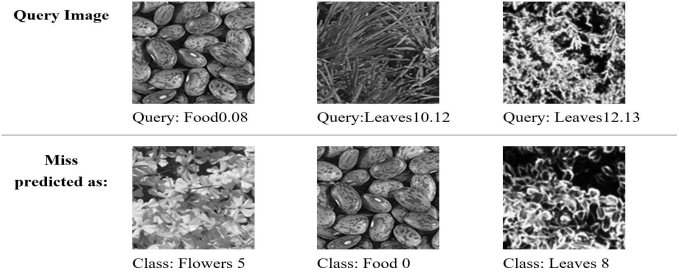

**Fig. 5.** Examples of kNN-CBIR retrieval from Vistex-40. For each submitted query (first row), the predicted class membership is false (second row).

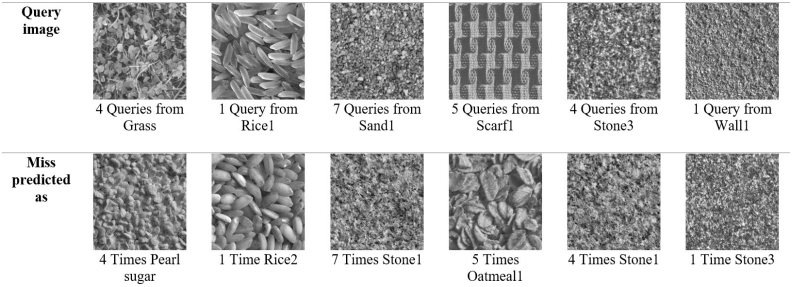

**Fig. 6.** Examples of SVM-CBIR retrieval from Kylberg-27 database. For each query (first row), the predicted class membership is false (second row).

**Table IV:** Performance comparison of ML-CBIR schemes vs. traditional CBIR schemes in terms of average retrieval rates (AR %) over 1080 images from kylberg-27 database.

| Feature extraction | *AR%* (kNN-CBIR) | *AR%* (Trad. CBIR) | Difference% |
|---|---|---|---|
| GGD1 | 97.69 | 65.79 | +31.90 |
| GGD2 | 97.31 | 64.95 | +32.36 |
| E | 93.43 | 63.77 | +29.66 |
| Feature extraction | *AR%* (SVM-CBIR) | *AR%* (Trad. CBIR) | Difference% |
| GGD1 | 97.96 | 65.79 | +32.17 |
| GGD2 | 97.59 | 64.95 | +32.64 |
| E | 93.43 | 63.77 | +29.66 |